\documentclass[conference]{IEEEtran}
\IEEEoverridecommandlockouts
\usepackage{cite}
\usepackage{amsmath,amssymb,amsfonts}
\usepackage{graphicx}
\usepackage{textcomp}
\usepackage{xcolor}

\usepackage{hyperref}       %
\usepackage{url}            %
\usepackage{booktabs}       %
\usepackage{amsfonts}       %
\usepackage{nicefrac}       %
\usepackage{microtype}      %
\usepackage{lipsum}		%
\usepackage{graphicx}
\usepackage{doi}

\def\BibTeX{{\rm B\kern-.05em{\sc i\kern-.025em b}\kern-.08em
    T\kern-.1667em\lower.7ex\hbox{E}\kern-.125emX}}
\usepackage[labelfont=bf,tableposition=top]{caption}
\usepackage{colortbl}

\usepackage{algorithm}
\usepackage{algpseudocode}

\usepackage{booktabs}       %
\usepackage{pifont}%
\newcommand{\cmark}{\ding{51}}%
\newcommand{\xmark}{\ding{55}}%

\usepackage{multirow}
\usepackage{amsfonts}       %
\usepackage{nicefrac}       %
\usepackage{microtype}      %
\usepackage{xcolor}         %
\usepackage{caption}
\usepackage{tikz}
\usepackage{tablefootnote}

\usepackage{amsthm}
\usepackage{amsmath}
\newtheorem{definition}{Definition}
\usepackage{subcaption}
\usepackage{makecell}

\definecolor{shadecolor}{rgb}{1,0.8,0.3}
\definecolor{lightgray}{rgb}{0.9,0.9,0.9}
\definecolor{yellow}{rgb}{1,0.5,0}
\definecolor{lightblue}{rgb}{0.7569, 0.8824, 0.9255}

\newcommand{\best}[1]{\textcolor{black}{\textbf{\underline{#1}}}}
\newcommand{\second}[1]{\textcolor{black}{\underline{#1}}}
\newcommand{\update}[1]{\textcolor{black}{{#1}}}

\def\BibTeX{{\rm B\kern-.05em{\sc i\kern-.025em b}\kern-.08em
    T\kern-.1667em\lower.7ex\hbox{E}\kern-.125emX}}
\begin{document}

\newcommand{\thename}{PARDON}

\title{
PARDON: \underline{P}rivacy-\underline{A}ware and \underline{R}obust \\Federated \underline{Do}main Generalizatio\underline{n}
\thanks{We acknowledge partial support from the National Security Agency under the Science of Security program, from the Defense Advanced Research Projects Agency under the CASTLE program, and from an Amazon Research Award.  The contents of this paper do not necessarily reflect the views of the US Government.}
}

\author{\IEEEauthorblockN{Dung Thuy Nguyen}
\IEEEauthorblockA{\textit{Department of Computer Science} \\
\textit{Vanderbilt University}\\
Nashville, TN, USA \\
dung.t.nguyen@vanderbilt.edu}
\and
\IEEEauthorblockN{Taylor T. Johnson}
\IEEEauthorblockA{\textit{Department of Computer Science} \\
\textit{Vanderbilt University}\\
Nashville, TN, USA \\
taylor.johnson@vanderbilt.edu}
\and
\IEEEauthorblockN{Kevin Leach}
\IEEEauthorblockA{\textit{Department of Computer Science} \\
\textit{Vanderbilt University}\\
Nashville, TN, USA \\
kevin.leach@vanderbilt.edu}
}

\maketitle

\begin{abstract}
While Federated Learning (FL) shows promise in preserving privacy and enabling collaborative learning, most current solutions concentrate on private data collected from a single domain. 
Yet, a substantial performance degradation on unseen domains arises when data among clients is drawn from diverse domains (i.e., domain shift).
However, existing Federated Domain Generalization (FedDG) methods are typically designed under the assumption that each client has access to the complete dataset of a single domain. This assumption hinders their performance in real-world FL scenarios, which are characterized by domain-based heterogeneity—where data from a single domain is distributed heterogeneously across clients—and client sampling, where only a subset of clients participate in each training round.

In addition, certain methods enable information sharing among clients, raising privacy concerns as this information could be used to reconstruct sensitive private data.
To overcome this limitation, we present \thename{}, a novel FedDG paradigm designed to robustly handle more complicated domain distributions between clients while ensuring security. 
\thename{} facilitates client learning across domains by extracting an interpolative style from abstracted local styles obtained from each client and using contrastive learning. 
This approach provides each client with a multi-domain representation and an unbiased convergent target.
Empirical results on multiple datasets, including PACS, Office-Home, and IWildCam, demonstrate \thename{}'s superiority over state-of-the-art methods. 
Notably, our method outperforms state-of-the-art techniques by a margin ranging from 3.64 to 57.22\% in terms of accuracy on unseen domains. 
Our code is available at \url{https://github.com/judydnguyen/PARDON-FedDG}.
\end{abstract}

\section{Introduction}
\label{sec:intro}

\textit{Federated learning }(FL)~\cite{mcmahan2017communication} is a distributed machine learning paradigm that facilitates training a single unified model from multiple disjoint contributors, each of whom may own or control their private data. 
The security aggregation mechanism and its distinctive distributed training mode render it highly compatible with a wide range of practical applications that have strict privacy demands~\cite{sheller2020federated,nguyen2021federated}.  
Data heterogeneity~\cite{li2020federated,karimireddy2020scaffold,mendieta2022local,lim2024metavers} presents a substantial challenge in FL due to the diverse origins of data, each with a distinct distribution.
In this scenario, each client optimizes towards the local minimum of empirical loss, deviating from the global direction. 
Therefore, the averaged global model unavoidably faces a slow convergence speed~\cite{li2020federated} and achieves limited performance improvement~\cite{Wang2020federated}. 
Many techniques have been developed to handle data heterogeneity in FL~\cite{gao2022federated,li2021model,nguyen2022feddrl}.
However, these techniques primarily address label skew --- i.e., the variation in label distributions among client's data within the same domain. 

\begin{figure}[t]
    \centering
    \includegraphics[width=1.0\linewidth]{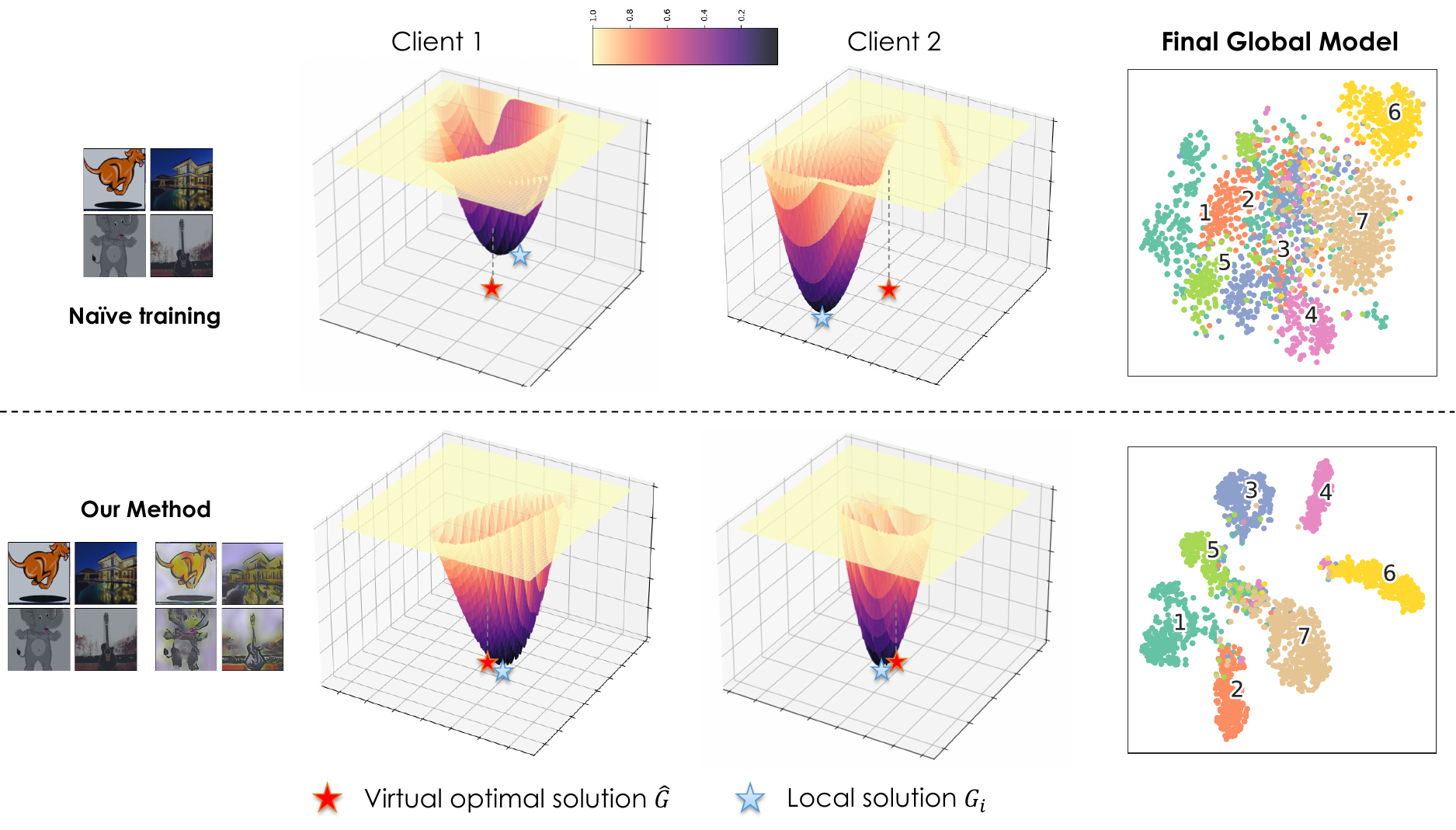}
    \caption{Loss landscape visualization of two clients under domain-based heterogeneity, with normal training (first row) and our method using interpolative style-transferred data (second row). The vertical axis shows the loss (i.e., solution for the global optimum $G$ and each local objective $G_i$). The horizontal plane represents a parameter space centered at the global model weight. The feature visualization in the last column demonstrates that our method achieves superior classification results on unseen domains.}
    \label{fig:fisc-intro}
    \vspace{-0.4cm}
\end{figure}
However, real-world data heterogeneity in practical applications extends beyond class imbalances, introducing complexities such as domain shifts due to geographical dispersion of data collectors, like cameras and sensors across different hospitals in medical settings~\cite{orlando2020refuge, guo2023domaindrop, huang2023rethinking}.
This problem is widely-addressed in centralized machine learning, where all training data is collected in central storage.
Most existing
DG methods for centralized ML~\cite{sun2016deep,nguyen2021domain,wang2021discriminative} require sharing the representation of the data across domains or access to data from a pool of multiple source domains (in a central server)~\cite{li2018learning}. 
However, this violates FL's privacy-preserving nature and raises a critical need for specialized DG methods in FL.

Recognizing this issue, several Federated Domain Generalization (FedDG) methods have attempted to address domain shift in FL~\cite{liu2021feddg,nguyen2022fedsr,tenison2023gradient,zhang2023federated}, but their methods and evaluations often reveal several limitations. 
Firstly, current evaluations are confined to testing on datasets with limited domain diversity and a small number of domains. 
Secondly, existing approaches are highly sensitive to the number of clients involved in training, and \textit{domain-based client heterogeneity}~\cite{bai2024benchmarking} where each client distribution is a (different) mixture of train domain distributions.  
 In addition, \textit{client sampling} scenarios where only a subset of clients engage are often overlooked, limiting the generalizability of these methods to popular FL setups~\cite{fu2023client,crawshaw2024federated}. \update{We argue that existing techniques rely on local signals, such as training loss and gradient sign, to mitigate overfitting by penalizing model complexity. However, by treating each client signal as a complete domain indicator, these methods may struggle to generalize when a single client lacks comprehensive domain data.} In addition, other methods using \update{cross-sharing local signals, such as amplitude, style statistics, and class-level prototypes, among clients can
 potentially result in privacy breaches, as discussed in prior studies~\cite{nguyen2022fedsr,zhang2023federated,liu2024perspective}.
} 

To address these challenges, we propose shifting from relying on individual local information in each round to using an aggregated \textit{interpolation style}, which fuses shared domain knowledge from all clients from the beginning to mitigate the limitations caused by \textit{client sampling}. This global \textit{interpolation style} is formed through two-level clustering and unbiased blending of local style statistics, creating a representative style that captures inter-client domain diversity to address the limitations of \textit{domain-based client heterogeneity}.
By sharing this global style instead of specific local signals,
we reduce privacy risks while maintaining superior performance, as the global style conceals client-specific data and class-level information.
Given the \textit{interpolation style}, each client can generate new data containing the global interpolative style, leveraging domain diversity from all clients. 
We enforce the model to represent the original data in a manner that is more similar to the style-transferred data from the same class while pushing it away from the style-transferred data from other classes. This approach enhances generalization by encouraging the model to focus on domain-invariant features, gradually aligning it with the global model and preventing bias towards local data. Fig.~\ref{fig:fisc-intro} intuitively illustrates local models converging towards their local optima via loss landscape visualization. Our method aims at guiding a local model to optimize towards a converged optimal solution for inter-domain data. This results in better discrimination of unseen domains, as illustrated in the TSNE plot in the rightmost column of Fig.~\ref{fig:fisc-intro}.

We claim the following major contributions in this work.
\begin{itemize}
    \item We are the first to investigate FedDG methods under both \textit{domain-based client heterogeneity} and \textit{client sampling} scenarios, pinpointing the underlying limitation of current methods: the client's signals in each round are insufficient for representing the complete global domain knowledge, resulting in a model distorted by partial observation.
    \item 
    We present \thename{}, a FedDG method that enhances robustness and privacy under client sampling and domain heterogeneity.  
    \thename{} employs multi-level unsupervised clustering to identify an unbiased interpolative style vector that represents styles across client domains.
     Then, we introduce a multi-domain contrastive learning mechanism that guides each local model to learn a multi-domain representation aligned with the global representation, reducing local bias from individual datasets.
    \item We rigorously validate the effectiveness of our resulting model with a wide range of datasets (i.e., both small domain and large domain),  varied domain distribution, and number of client settings. Our results demonstrate that \thename{} outperforms existing FedDG methods, maintaining comparable computational overhead while effectively preserving client privacy and sensitive information. 
\end{itemize}

\section{Related Work}

While FedDG shares a common objective with centralized Domain Generalization (DG), i.e., generalizing from multi-source domains to unseen domains, FedDG is distinct from traditional DG in prohibiting direct data sharing among clients. 
This section first examines how domain generalization is tackled in centralized ML approaches, then delves into existing FedDG methodologies closely related to ours.

\noindent\textbf{Domain Generalization.} 
Domain generalization aims to learn a model from multiple source domains so that the model can generalize to an unseen target domain.  Existing work has explored domain shifts from various directions in a centralized data setting. These methods can be divided into three categories~\cite{wang2022generalizing}, including \textit{data manipulation} to enrich data diversity~\cite{shankar2018generalizing,zhou2021domain}, \textit{domain-invariant feature} distribution across domains for the robustness of conditional distribution to enhance the generalization ability of the model~\cite{nguyen2021domain,mitrovic2021representation}, and exploiting general learning strategies such as \textit{meta-learning-based}~\cite{niu2023knowledge} and transfer learning~\cite{zhang2022fine} to promote generalizability. 
However, many of these methods require centralized data from different domains, violating local data preservation in federated learning. Specifically, access for more than one domain is needed to augment data or generate new data in~\cite{shi2022gradient}; domain-invariant representation learning or decomposing features are performed under the comparison
across domains, and some learning strategy-based
methods use an extra domain for meta-update~\cite{arjovsky2019invariant,piratla2020efficient}. 
There exists some methods that do not explicitly require centralized domains and can be adapted for federated learning with minor adjustments.
However, these methods such as MixStyle~\cite{zhou2021domain},\cite{xu2021fourier} and JiGen~\cite{carlucci2019domain}  offer minimal improvement in addressing domain shift in FL due to constrained intra-client and differing inter-client distributions, as noted by~\cite{chen2023federated, bai2024benchmarking}. This highlights the preference for a specialized FedDG method that preserves the distributed nature of FL while efficiently enhancing generalizability.

\noindent\textbf{Federated Domain Generalization. }
Existing works tackling heterogeneity among clients in FL~\cite{hanzely2020lower,karimireddy2020scaffold,Qu_2022_CVPR} have not considered model performance under the domain shift between training and testing data. Recently, works tackling DG in FL~\cite{liu2021feddg,nguyen2022fedsr,huang2023rethinking,zhang2023federated,tenison2023gradient,chen2023federated} have been proposed. First, \textit{cross-information sharing} lets other clients see information abstracted from one client's domain --- for example, the amplitude spectrum~\cite{liu2021feddg,shenaj2023learning} and the representation statistics~\cite{park2024stablefdg,chen2023federated}. Other clients then use this information to produce synthetic data that contains domain information to enrich the local data. In these methods, clients in FL are encouraged to share their sample-level or class-level information. 
Thus, these methods cause additional costs and risks of data privacy leakage since the style information of each client is publicly available to other clients or the server.  Second, \textit{multi-object optimization} aligns the generalization gaps among clients. For example, Zhang et al.~\cite{zhang2023federated} create a new optimization goal with a regularizer to lower variance and get a tighter generalization bound. Nguyen et al.~\cite{nguyen2022fedsr} proposed restricting the representation's complexity to help minimize the distribution distance between specific and reference domains. In Tenison et al.~\cite{tenison2023gradient}, the authors suggested using a gradient-masked averaging aggregation method based on the signed agreement of the local updates to eliminate weight parameters with much disagreement.
Besides these two approaches, 
third, \textit{unbiased learning}~\cite{huang2023rethinking} improves domain generalization by creating unbiased by-class prototypes and using regularization to line up the local models with the unbiased prototypes. 
However, previous studies on FedDG have often focused on domain-isolated settings where the number of clients equals the number of training domains. In addition, these methods show limited effectiveness in more complicated scenarios such as domain-based heterogeneity and datasets with a large number of domains~\cite{bai2024benchmarking}. Moreover, client sampling, a common setting in FL~\cite{fu2023client,crawshaw2024federated}, is typically overlooked, where only a portion of clients participate in each training round. Considering client sampling is essential as it reflects real-world scenarios where involving all clients in every round is impractical due to constraints such as communication bandwidth, computational resources, and energy limitations~\cite{karimireddy2020scaffold,fu2023client}. 
Unlike prior methods, we relax the assumption of isolated domain distribution by considering various client-domain heterogeneity and client sampling scenarios. 
We argue that efficient FedDG methods must perform robustly and securely under these settings with reasonable computation cost.

\label{sec:related}

\section{Methodology}
\label{sec:method}
\begin{figure*}[t]
\vspace{-0.2cm}
    \centering
\includegraphics[width=0.9\textwidth]{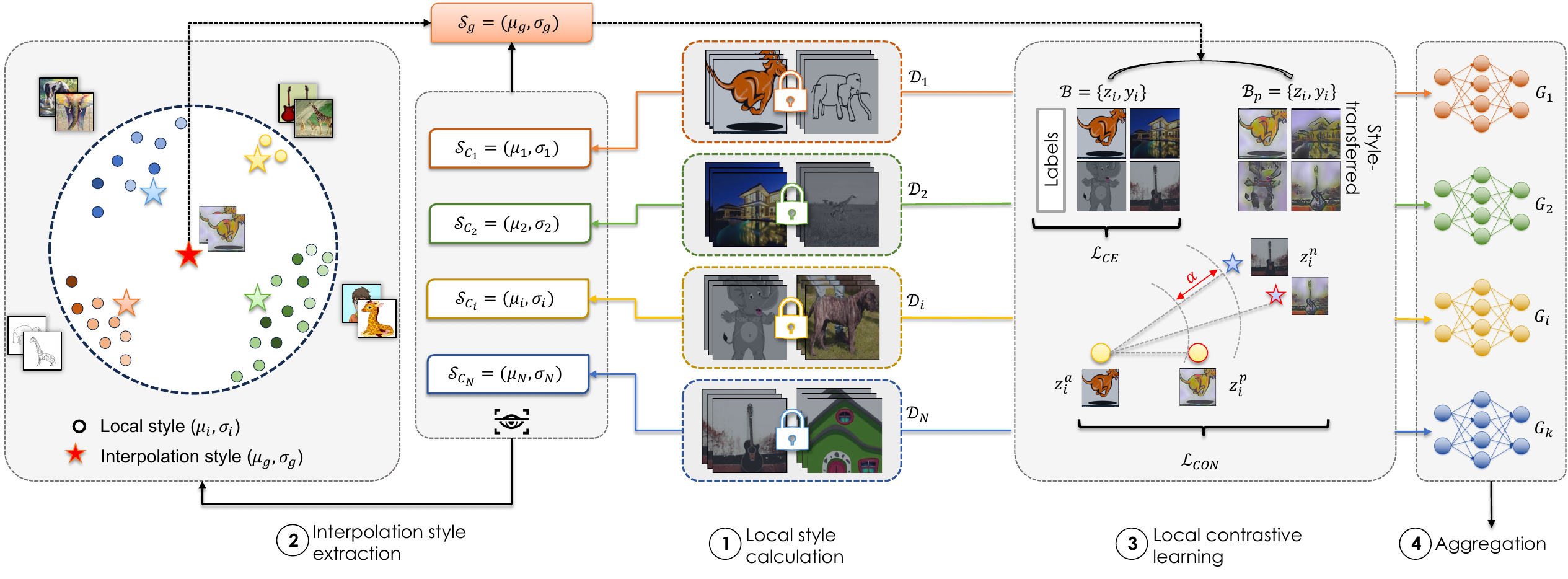}
\vspace{-0.2cm}
    \caption{The \thename{} framework we present in this paper.
    Clients calculate their style information using their local data and a pre-trained encoder in Step~\textcircled{1}. The server extracts interpolation style by weighted averaging the local style information in Step~\textcircled{2}. Then, in Step~\textcircled{3}, local clients exploit the interpolation style to obtain style-transferred data and update their models via contrastive learning. The aggregated global model in Step~\textcircled{4} runs inference on unseen data.
    }
    \label{fig:fedtrans-overall}
\end{figure*}
In this section, we formally introduce the domain-related definitions used in this work and then present \thename{}, our approach for domain generalization in FL.

\subsection{Preliminaries}
First, we extend domain and domain generalization definitions in~\cite{wang2022generalizing} into domain shift and domain-based client heterogeneity~\cite{bai2024benchmarking} within the context of FL.
\begin{definition}[Domain]
    Let $\mathcal{X}$ denote a nonempty input space and $\mathcal{Y}$ an output space. A \emph{domain} comprises data sampled from a distribution. We denote it as $\mathcal{S}=\left\{\left(\mathbf{x}_i, y_i\right)\right\}_{i=1}^n \sim$ $P_{X Y}$, where $\mathbf{x} \in \mathcal{X} \subset \mathbb{R}^d, y \in \mathcal{Y} \subset \mathbb{R}$ denotes the label, and $P_{X Y}$ denotes the joint distribution of the input sample and output label. $X$ and $Y$ denote the corresponding random variables.
\end{definition}

\begin{definition}[Domain Generalization]
We are given $M$ training (source) domains $\mathcal{S}_{\text {train }}=\left\{\mathcal{S}^i \mid i=1, \cdots, M\right\}$ where $\mathcal{S}^i=\left\{\left(\mathbf{x}_j^i, y_j^i\right)\right\}_{j=1}^{n_i}$ denotes the $i$-th domain. The joint distributions between each pair of domains are different: $P_{X Y}^i \neq P_{X Y}^j, 1 \leq i \neq j \leq M$. \\
The goal of \emph{domain generalization} is to learn a robust and generalizable predictive function $h: \mathcal{X} \rightarrow \mathcal{Y}$ from the $M$ training domains to achieve a minimum prediction error on an unseen test domain $\mathcal{S}_{\text {test }}$ (i.e., $\mathcal{S}_{\text {test }}$ cannot be accessed in training and $P_{X Y}^{t e s t} \neq P_{X Y}^i$ for $\left.i \in\{1, \cdots, M\}\right)$, i.e., 
$
\min _h \mathbb{E}_{(\mathbf{x}, y) \in \mathcal{S}_{\text {test }}}[\ell(h(\mathbf{x}), y)],
$
where $\mathbb{E}$ is the expectation and $\ell(\cdot, \cdot)$ is the loss function.
\end{definition}
\begin{definition}[Domain Shift in FL]
We consider a scenario where $M$ training domains $S_{\text{train}}$ are distributed among $K$ participants (indexed by $k$) with respective private datasets $\mathcal{D}_k = \left\{ (\mathbf{x}_i^k, y_i^k) \right\}_{i=1}^{N_k}$, where $N_k$ denotes the local data scale of participant $k$. In heterogeneous FL, the conditional feature distribution $P(\mathbf{x}|y)$ may vary across participants, even if $P(y)$ remains consistent, leading to domain shift: 
$P_k(\mathbf{x} \mid y) \neq P_l(\mathbf{x} \mid y), \text{ where } P_k(y)=P_l(y).$
\end{definition}

\begin{definition}[Domain-based Client Heterogeneity]
\label{def:domain-based-heter}
We consider domain-based client heterogeneity as an instance of domain shift in FL, such that
each client distribution $\mathcal{D}_i$ is a mixture of training domain distributions, which is: $\mathcal{D}_i(\mathbf{x}, y)=\sum_{d \in \mathcal{S}_{\text{train}}} w_{i, d} \cdot S_d(\mathbf{x}, y),$
where $w_{i, d}$ represents the proportion of samples from domain $d$ within client $i$.
\end{definition}

\subsection{\thename{}: Architecture}
Fig.~\ref{fig:fedtrans-overall} presents the overall architecture of \thename{}. 
Our method includes four main processes:~\textcircled{1}~local style calculation,~\textcircled{2} interpolation style extraction,~\textcircled{3} contrastive-based local training, and,~\textcircled{4}~aggregation. In this subsection, we present how each step is performed.
In this work, we adopt the FL scheme formalized from previous works~\cite{mcmahan2017communication,karimireddy2020scaffold}, in which all clients agree on sharing a model with the same architecture. We regard the model as having two modules: a feature extractor and a unified classifier. The feature extractor $f: \mathcal{X} \rightarrow \mathcal{Z}$, encodes sample $x$ into a compact $d$ dimensional feature vector $z=f(x) \in \mathbb{R}^d$ in the feature space $\mathcal{Z}$.
A unified classifier $g: \mathcal{Z} \rightarrow \mathbb{R}^{|I|}$ maps feature $z$ into logits output $y=g(z)$,
where $I$ is the classification category. 

\noindent\textbf{Local style calculation. }Initially, each client must compute their unique style and upload it to the central server. The style information extracted from each local client is carefully abstracted to ensure that it cannot be employed to reconstruct the original dataset. For this purpose, we have chosen to represent the style of each client by statistics of its local data's feature embedding. Importantly, it has been demonstrated to be highly challenging to reverse-engineer the original dataset solely from this style information (Chen et al.~\cite{chen2023federated}). 
Firstly, to extract the feature representation of each image, we use a pre-trained VGG encoder $\Phi$ of real-time style transfer model AdaIN~\cite{huang2017arbitrary}, which will be used to generate interpolative style-transferred data at the next step. 

Assume the feature $\Phi (x)$ has the shape of $d \times h \times w$ given channel dimension $x$, in which $h \times w$ is the feature map resolution, and $d$ is the number of dimensions. Then, denote the representation set of a client $\mathcal{C}_k$ as $\Phi(\mathcal{C}_k)$, where $ \Phi(\mathcal{C}_k) := \{\Phi(x_i) | \ \forall x_i \in \mathcal{D}_k\}$.
Since each client $k$ may contain data from more than one domain, we cannot estimate precisely the ratio of each domain, i.e., $w_{i, d}$ in Definition~\ref{def:domain-based-heter}. Still, we must prevent the local style from being biased by the dominant domains. 
Therefore, we categorize samples based on different styles using FINCH~\cite{sarfraz2019efficient}. This hierarchical clustering method identifies groupings in the data based on its neighborhood relationship without needing hyper-parameters or thresholds. It enables us to discover inherent clustering tendencies inside each client's samples. 
Compared to traditional clustering methods, FINCH is more suitable for scenarios where each client contains an uncertain number of clusters. %
Mathematically, given a set of features $\mathcal{F}$, FINCH outputs a set of partitions $\mathcal{L}=\left\{\Gamma_1, \Gamma_2, \cdots, \Gamma_L\right\}$ where each partition 
$\Gamma_i = \{C_1, C_2, \cdots, C_{\Gamma_i} | \ C_{\Gamma_i} > C_{\Gamma_{i+1}} \forall i \in \mathcal{L}\}$ is a valid clustering of $\mathcal{F}$, and the last clustering $\Gamma_L$ has the smallest number of classes.
Here, we use FINCH to group local data points based on their representation $\Phi(\mathcal{C}_k)$. Specifically,
\begin{equation}
\Phi(\mathcal{C}_k) \xrightarrow{\text { FINCH }} \Gamma_{L_k} = \left\{\epsilon_j \subset \Phi(\mathcal{C}_k) \mid \epsilon_j \cap \epsilon_k = \emptyset, \ \forall j \neq k\right\}_{j=1}^{L_k}
\end{equation}
This technique implicitly clusters samples with similar characteristic styles, as styles from other domains are less likely to be adjacent. Consequently, samples from distinct domains are unlikely to merge, while those from similar domains are naturally grouped. Specifically, we employ cosine similarity to quantify the closeness between two image styles, identifying the style with the smallest distance as its ``neighbor.''

Given a cluster \(\epsilon_j\), the concatenation of all its elements, \(\Phi(\epsilon_j)\), has the shape \(\mathbb{R}^{|\epsilon_j| \times d \times w \times h}\). The style of the cluster, denoted as \(\mathcal{S}(\epsilon_j)\), is derived as the pixel-level channel-wise mean and standard deviation of the feature representations across all elements. Formally, \(\mathcal{S}(\epsilon_j)\) is computed as follows:
\begin{equation}
\mathcal{S}(\epsilon_j) =
\left(\mu\left(\Phi(\epsilon_j)\right), \sigma\left(\Phi(\epsilon_j)\right)\right)\in \mathbb{R}^{2d} 
\end{equation}
The local style information for each client $k$ is now the set of all clusters' styles, i.e., 
$\mathcal{S}_{\mathcal{C}_k} := 
\{\mathcal{S}(\epsilon_j)\}_{j=1}^{L_k}$ and the client's style statistic is then determined by the average of all cluster styles, i.e., 
$S_{\mathcal{C}_k} = \frac{1}{L_k}\sum_{j=1}^{L_k} \mathcal{S}(\epsilon_j) \in \mathbb{R}^{2d}$.

\noindent\textbf{Interpolation Style Extraction. }
Since we consider FL settings with many participants and domain-based client heterogeneity among them, given a set of local styles 
\(\mathbf{S} = \left\{\mathcal{S}_{\mathcal{C}_i}\right\}_{i=1}^{N}\), our objective is to find an optimal combination of these local styles, denoted as \(\mathcal{S}_g\). Specifically, we employ FINCH to identify grouping relationships among client styles, forming clusters that capture shared characteristics and account for the inherent heterogeneity among participants.
\begin{equation}
\mathbf{S} \xrightarrow{\text{FINCH}} \Gamma_{L} = \left\{\varepsilon_j \subset \mathbf{S} \mid \varepsilon_j \cap \varepsilon_k = \emptyset, \ \forall j \neq k \right\}_{j=1}^{L}
\label{eq:finch-global}
\end{equation}
The output of this process is a set of \(L\) non-overlapping subsets of client styles from \(\mathbf{S}\). Since different clients may share similar domains or multiple clients may belong to the same domain, we reduce the number of local style representations to \(L\) cluster styles, as described in Eq.~\ref{eq:finch-global}. Each cluster \(j\) is represented by the average style of all \(S_{\mathcal{C}_i} \in \varepsilon_j\), computed as:
\begin{equation}
    \mathcal{S}(\varepsilon_j) = \frac{1}{|\varepsilon_j|} \sum_{S_{\mathcal{C}_i} \in \varepsilon_j} S_{\mathcal{C}_i} \in \mathbb{R}^{2d}
\end{equation}

After this step, the clustering method groups clients with close styles together, and we consider each cluster style equally rather than treating each client equally. Finally, the global interpolation style $\mathcal{S}_g$ is defined as the median of all cluster style statistics, i.e., 
\begin{equation}
\mathcal{S}_g = median\left(\mathcal{S}(\varepsilon_j) \ | \ \forall \varepsilon_j \in \Gamma_{L}\right) \in \mathbb{R}^{2d}
\end{equation}
This enables domains with low cardinality to engage in the interpolation style, hence facilitating the dominant domain's knowledge acquisition from these smaller populations. 
The median is robust to outliers (i.e., extreme values) and skewed distributions in style statistics, ensuring no single dominant style skews the global interpolation style. This fairness in representing the central tendency makes the median ideal for capturing diverse client styles without bias, promoting equitable and comprehensive knowledge transfer across all domains.
This strategy for calculating the \textit{interpolation style} effectively addresses the bias introduced by domain-based client heterogeneity. By incorporating information from all clients prior to the commencement of training, it circumvents the limitations associated with partial observations due to client sampling during individual training rounds, thereby enhancing the performance compared to other methods reliant on round-specific client participation.

\noindent\textbf{Local Contrastive Learning. }This component demonstrates how each client leverages the global interpolation style, $\mathcal{S}_g$, sent from the server, to learn an unbiased local model $G_i$. 
Motivated by the success of contrastive learning in FL, we introduce a multi-domain contrastive learning mechanism specialized for our method, which balances between \textit{inter- and intra-client domain diversity learning}.
The key idea is to use the feature representation of style-transferred data, with the global interpolation style as positive anchors for each data point in $\mathcal{D}_i$. 
Through a contrastive manner, the local client learns a feature extractor that aligns its local data and the style-transferred data. 
Specifically, this learning mechanism improves generalizability by treating samples from the same class with original and transferred styles as positive pairs and samples from different classes with both styles as negative pairs. 
The global interpolative style, denoted as $\mathcal{S}_g$, serves as an unbiased aggregation of the statistical information across all client domains. By encompassing multi-domain characteristics, $\mathcal{S}_g$ provides a balanced and inclusive representation that acts as a guiding framework for achieving a fair and convergent learning objective. This strategy enhances the model's ability to capture more refined and discriminative class-level features while simultaneously encouraging the learning of domain-invariant features. As a result, it improves generalization performance and mitigates biases caused by domain heterogeneity.
To achieve style-transferred data $\mathcal{D}_i'$ given $\mathcal{D}_i$ and $\mathcal{S}_g$, we use pre-trained AdaIN model, in which:
\begin{equation}
\operatorname{AdaIN}\left(F_{\mathcal{D}_i}, \mathcal{S}_g\right)=\sigma\left(\mathcal{S}_g\right)\left(\frac{f({\mathcal{D}_i})-\mu\left(f({\mathcal{D}_i})\right)}{\sigma\left(f({\mathcal{D}_i})\right)}\right)+\mu\left(\mathcal{S}_g\right),
\end{equation}
where $f({\mathcal{D}_i})$ is the representation of $\mathcal{D}_i$ in feature space, and
$\mu(.)$ and $\sigma (.)$ are the channel-wise mean and standard variance of image features, respectively.
Given a training batch $\mathcal{B} = \{z_i, y_i\}$ with $z_i, y_i$ are the feature representation and label of $i^{th}$ sample, respectively, we denote the corresponding style-transferred batch as $\mathcal{B}_p$ such that 
$
    \mathcal{B}_p = \{\operatorname{AdaIN}\left(z_i, \mathcal{S}_g\right), y_i | \ \forall z_i \in \mathcal{B}\}
$.

To leverage this style-transferred data, we use triplet loss \cite{schroff2015facenet} to guide clients to learn from embedding following a globally agreed style. In action,
for each training batch,~\thename{} would treat all representations of the ground-truth samples as the set of anchors and
build up their corresponding positive and negative sets. Specifically, the positive sample for anchor $z_i$ is defined as its corresponding style-transferred embedding, i.e., $z_i^p := z'_i \in \mathcal{B}_p$, whereas the negative sample is style-transferred embedding from other classes. Consider the set of negative anchors as: 
$
\mathcal{N}_i = \{z_j|z_j \in \mathcal{B}_p \wedge y_j \neq y_i\}.
$
Then, 
the triplet loss function is expressed as follows:

\begin{equation}
{\mathcal{L}_T }=\sum_{i=1}^{|\mathcal{B}|}\left[\left\|z_i^a-z_i^p\right\|_2^2 - \frac{1}{|\mathcal{N}_i|} \sum_{z_i^n \in \mathcal{N}_i} \left\|z_i^a-z_i^n\right\|_2^2 + \alpha \right],
\end{equation}
Here, $z_i^a, z_i^p, z_i^n$ are feature representations of the current sample $x_i$, its positive and negative anchors, respectively, and $\alpha$ is the margin value.

\textit{Intra-client domain diversity learning. }
In addition to distinguishing inter-domain representations, each client should have the capability to effectively learn from its local data, i.e., intra-client domain diversity. To achieve this, we incorporate the CrossEntropy~\cite{de2005tutorial} loss, utilizing the logits output along with the original annotation signal \(\left(y_i\right)\) to maintain local domain discriminatory power. This is formulated as:
\(
\mathcal{L}_{CE} = -\mathbf{1}_{y_i} \log \left(\sigma\left(g\left(z_i\right)\right)\right),
\)
where \(\sigma\) denotes the softmax function. Since the cross-entropy loss term already encourages the model to leverage intra-client domain diversity, we use an additional mechanism to ensure the model can effectively learn from data across various domains without overfitting its local data. To address this, we impose restrictions on the amount of information the model can encode by applying L2 regularization directly to the representations, rather than the network parameters. This approach limits the complexity of the learned embeddings and enhances their generalizability~\cite{xu2016new,nguyen2022fedsr}.

Specifically, we incorporate L2 regularization, denoted as \(\mathcal{L}_{reg}\), to prevent overfitting and encourage bounded embeddings in contrastive learning. The regularization term is defined as:

\begin{equation}
\mathcal{L}_{reg} = \sum_{i=1}^{|\mathcal{B}|} \|z_i\|_2^2 + \|z_i^p\|_2^2,
\label{eqn:regu}
\end{equation}
where \(\mathcal{B}\) represents the batch, and \(z_i\) is the representation of the \(i\)-th sample. This regularization encourages embeddings to remain within a bounded range, thereby improving the model's ability to generalize across diverse domains.

Finally, each participant \(C_{i}\) optimizes its local data during local updates by minimizing the following objective:

\begin{equation}
\mathcal{L} = \mathcal{L}_{CE} + \gamma_1 \mathcal{L}_{T} + \gamma_2 \mathcal{L}_{reg},
\label{eqn:loss_main}
\end{equation}
where \(\gamma_1\) and \(\gamma_2\) are coefficient terms that control the strength of the inter-domain learning and intra-domain regularization.

\noindent\textbf{Aggregation. }We adopt the original average aggregator from most FL frameworks, in which the global model $G$ is calculated by $G = \frac{1}{N}\sum_{i=1}^K n_iG_i$, where $n_i = |\mathcal{D}_i|$ is the data size of client $\mathcal{C}_i$ and $N = \sum_i^K n_i$.

To this end, we hypothesize that \thename{} can achieve state-of-the-art generalizability in complex and practical scenarios with domain-based client heterogeneity and client sampling, while preserving local data privacy.

\section{Experiments}
\label{sec:expriments}
In this section, we compare our method with five FedDG baselines under FL settings concerning varied training domains, number of clients, and domain-based client heterogeneity, then demonstrate \thename{} can achieve improved generalizability and security in complicated scenarios involving domain-based client heterogeneity and client sampling.
We conduct all the experiments using PyTorch 2.1.0~\cite{pytorch} and the framework provided by Bai et al.~\cite{bai2024benchmarking} to simulate domain-based client heterogeneity. 
All experiments are run on a computer with an Intel Xeon Gold 6330N CPU and an  NVIDIA A6000 GPU.
\subsection{Experimental Setup}
\begin{table*}[hbt]
\centering\scriptsize
\renewcommand{\arraystretch}{0.9}
\caption{Comparison of inter-domain performance with SOTA methods under different split LTDO schemes, i.e., two domains are used for validation. AVG denotes the average accuracy calculated from each domain. (The best average
accuracy is marked in \best{bold-underline}. The second-best average accuracy is marked in \second{underline}.)}
\label{tab:ltdo_full}
\resizebox{0.89\textwidth}{!}{
\begin{tabular}{@{}llccccc|ccccc@{}}
\toprule    
& 
& \multicolumn{5}{c}{Validation Accuracy}              
& \multicolumn{5}{c}{Test Accuracy}                          \\ \cmidrule(lr){3-7} \cmidrule(lr){8-12} 
\multirow{1}{*}{Dataset}  
& \multirow{1}{*}{Methods}                        
& \multicolumn{1}{c}{A} 
& \multicolumn{1}{c}{P} 
& \multicolumn{1}{c}{C} 
& \multicolumn{1}{c}{S}             
& \multicolumn{1}{c}{AVG}   
& P                         
& S                         
& A                         
& \multicolumn{1}{c}{C}     
& \multicolumn{1}{c}{AVG}   \\ \cmidrule(lr){3-12}
\multicolumn{1}{c|}{\multirow{5}{*}{\rotatebox{90}{PACS}}}       
& FedSR                    
& 14.80\%                  
& 14.67\%                 
& 13.39\%                 
& \multicolumn{1}{c|}{13.36\%}         
& \multicolumn{1}{c|}{\cellcolor{lightblue}14.06\%} 
& 13.80\%                      
& 13.97\%                     
& 14.55\%                     
& \multicolumn{1}{c|}{12.87\%} 
& \multicolumn{1}{c|}{\cellcolor{lightblue}13.80\%} \\
\multicolumn{1}{c|}{}                            
& FedGMA                   
& 39.31\%                 
& 94.13\%                 
& 63.95\%                 
& \multicolumn{1}{c|}{36.22\%}         
& \multicolumn{1}{c|}{\cellcolor{lightblue}58.40\%} 
& 73.83\%                     
& 64.85\%                     
& 73.10\%                     
& \multicolumn{1}{c|}{52.73\%} 
& \multicolumn{1}{c|}{\cellcolor{lightblue}66.13\%} \\
\multicolumn{1}{c|}{}                            
& FPL                      
& 77.93\%                 
& 94.49\%                 
& 64.97\%        
& \multicolumn{1}{c|}{31.61\%}         
& \second{\cellcolor{lightblue}67.25\%}                      
& 93.53\%                     
& 55.97\%                     
& 62.01\%                     
& \multicolumn{1}{c|}{51.83\%} 
& \multicolumn{1}{c|}{\cellcolor{lightblue}65.84\%} \\
\multicolumn{1}{c|}{}                            
& FedDG-GA                  
& 64.99\%                 
& 92.46\%                 
& 63.18\%                 
& \multicolumn{1}{c|}{32.73\%}         
& \cellcolor{lightblue}63.34\%                      
& 84.19\%                     
& 63.55\%                     
& 61.87\%                     
& \multicolumn{1}{c|}{48.08\%} 
& \multicolumn{1}{c|}{\cellcolor{lightblue}64.42\%} \\
\multicolumn{1}{c|}{} 

& CCST
& 68.51\%
& 96.41\%
& 59.26\%                 
& \multicolumn{1}{c|}{35.68\%}         
& \cellcolor{lightblue}64.97\%                      
& \multicolumn{1}{r}{86.89\%} 
& \multicolumn{1}{r}{59.91\%} 
& \multicolumn{1}{r}{71.78\%} 
& \multicolumn{1}{c|}{50.94\%} 
& \multicolumn{1}{c|}{\cellcolor{lightblue}\second{67.38\%}} \\
\cmidrule(lr){2-12}
\multicolumn{1}{c|}{}   

& Ours                     
& 73.63\%                 
& 95.57\%                 
& 69.41\%                 
& \multicolumn{1}{c|}{35.91\%}         
& \multicolumn{1}{c|}{\cellcolor{lightblue}\best{68.63\%}} 
& 93.05\%                     
& 66.20\%                     
& 71.73\%                     
& \multicolumn{1}{c|}{53.11\%} 
& \multicolumn{1}{c|}{\cellcolor{lightblue}\best{71.02\%}} \\ \midrule
&                          
& \multicolumn{1}{c}{C} 
& \multicolumn{1}{c}{A} 
& \multicolumn{1}{c}{R} 
& \multicolumn{1}{c}{P}             
& \multicolumn{1}{c}{AVG}   
& A   
& P  
& C                                                
& \multicolumn{1}{c}{R}     
& \multicolumn{1}{c}{AVG}   
\\ \cmidrule(l){3-12} 
\multicolumn{1}{c|}{\multirow{5}{*}{\rotatebox{90}{OfficeHome}}}                                
& FedSR                    
& 1.40\% 
& 1.24\% 
& 1.36\%                          
& \multicolumn{1}{c|}{1.31\%}
& \cellcolor{lightblue}1.33\%                       
& \multicolumn{1}{c}{1.15\%}
& \multicolumn{1}{c}{1.14\%}  
& \multicolumn{1}{c}{1.34\%}  
& \multicolumn{1}{c|}{1.33\%}  
& \multicolumn{1}{c|}{\cellcolor{lightblue}1.24\%}  \\
\multicolumn{1}{c|}{}                            
& FedGMA                   
& 43.18\%  
& 54.92\%    
& 66.81\%                     
& \multicolumn{1}{c|}{54.29\%}         
& \cellcolor{lightblue}\second{54.80\%}                       
& \multicolumn{1}{r}{55.71\%} 
& \multicolumn{1}{r}{66.43\%} 
& \multicolumn{1}{r}{39.91\%} 
& \multicolumn{1}{c|}{56.83\%} 
& \multicolumn{1}{c|}{\cellcolor{lightblue}54.72\%} \\
\multicolumn{1}{c|}{}                            
& FPL                      
& 45.72\%    
& 56.82\%    
& 69.45\%                 
             
& \multicolumn{1}{c|}{46.18\%}         
& \cellcolor{lightblue}54.54\%                      
& \multicolumn{1}{r}{59.95\%} 
& \multicolumn{1}{r}{65.22\%} 
& \multicolumn{1}{r}{43.99\%} 

& \multicolumn{1}{c|}{52.54\%} 
& \multicolumn{1}{c|}{\cellcolor{lightblue}\second{55.43\%}} \\
\multicolumn{1}{c|}{}                            
& FedDG-GA                  
& 38.99\%   
& 51.38\%   
& 63.85\%                 
             
& \multicolumn{1}{c|}{48.07\%}         
& \cellcolor{lightblue}50.57\%                      
& \multicolumn{1}{r}{51.63\%} 
& \multicolumn{1}{r}{62.38\%} 
& \multicolumn{1}{r}{36.68\%} 
& \multicolumn{1}{c|}{54.79\%} 
& \multicolumn{1}{c|}{\cellcolor{lightblue}51.37\%} \\
\multicolumn{1}{c|}{}      

& CCST                 
& 44.81\% 
& 52.48\%
& 62.29\%              
             
& \multicolumn{1}{c|}{49.85\%}         
& \cellcolor{lightblue}52.36\%                      
& \multicolumn{1}{r}{52.20\%} 
& \multicolumn{1}{r}{62.79\%} 
& \multicolumn{1}{r}{38.37\%}
& \multicolumn{1}{c|}{54.88\%} 
& \multicolumn{1}{c|}{\cellcolor{lightblue}52.06\%} \\
\cmidrule(lr){2-12}
\multicolumn{1}{c|}{}   
& Ours                     
& 46.74\%  
& 58.84\%   
& 71.13\%                 
              
& \multicolumn{1}{c|}{55.31\%}     
& \best{\cellcolor{lightblue}58.01\%}                      
& \multicolumn{1}{r}{60.09\%} 
& \multicolumn{1}{r}{67.54\%} 
& \multicolumn{1}{r}{45.41\%} 
& \multicolumn{1}{c|}{61.62\%} 
& \multicolumn{1}{c|}{\cellcolor{lightblue}\best{58.67\%}} \\ 
\bottomrule
\vspace{0.2cm}
& \multicolumn{8}{l}{PACS: \textit{A: Art-Painting, P: Photo, C: Cartoon, S: Sketch}} \\
& \multicolumn{8}{l}{OfficeHome: \textit{C: Clipart, A: Art, R: Real World, P: Product}} \\
\end{tabular}
}
\end{table*}

\begin{table*}[tbh]
\centering
\renewcommand{\arraystretch}{0.9}
\centering\scriptsize
\caption{Performance (\%) comparisons with the start-of-the-art
FedDG methods with the PACS dataset under LODO schemes, i.e., one domain is left for validation.}
\label{tab:lodo-full}
\resizebox{0.85\textwidth}{!}{
\begin{tabular}{@{}lcccccccccc@{}}
\toprule
\multirow{2}{*}{Methods} 
& \multicolumn{5}{c}{PACS}                                             & \multicolumn{5}{c}{OfficeHome}                                       \\ \cmidrule(lr){2-6} \cmidrule(lr){7-11}
& \multicolumn{1}{c}{P} 
& \multicolumn{1}{c}{A} 
& \multicolumn{1}{c}{C} 
& \multicolumn{1}{c}{S} 
& AVG & \multicolumn{1}{c}{P} 
& \multicolumn{1}{c}{A} 
& \multicolumn{1}{c}{C} 
& \multicolumn{1}{c}{R} 
& AVG \\ \midrule
FedSR
&   14.01\%                     
&   13.27\%                       
&   15.66\%                       
&   13.49\%                         
&   \multicolumn{1}{c|}{\cellcolor{lightblue}14.11\%}     
&   1.14\%                    
&   1.27\%                       
&   1.34\%                       
&   1.68\%                       
&   \cellcolor{lightblue} 1.36\% \\ 
FedGMA
&   91.08\%                    
&   81.25\%                       
&   66.04\%                    
&   61.72\%                          
&   \multicolumn{1}{c|}{\cellcolor{lightblue}75.02\%}     
&   69.99\%                       
&   61.60\%                       
&   49.87\%                       
&   71.93\%                       
&   \cellcolor{lightblue}63.35\%\\ 
FPL             
&   98.20\%                       
&   81.98\%                       
&   69.41\%                       
&   62.64\%                          
&   \multicolumn{1}{c|}{\cellcolor{lightblue}\second{78.06\%}}   
&   68.98\%                       
&   63.00\%                       
&   48.98\%                      
&   71.24\%                       
&   \cellcolor{lightblue}\second{63.05\%}\\
FedDG-GA
&   97.19\%                   
&   76.17\%                       
&   58.67\%                          
&   53.42\%                           
&   \multicolumn{1}{c|}{\cellcolor{lightblue}71.36\%}     
&   66.03\%                       
&   41.49\%                       
&   45.98\%                       
&   64.22\%                       
&   \cellcolor{lightblue}54.34\% \\ 
CCST   
& 97.07\%                    
& 84.18\%                   
& 72.44\%                          
& 59.43\%  
& \multicolumn{1}{c|}{\cellcolor{lightblue}78.28\%}        
& 66.93\%                   
& 56.82\%                   
& 47.35\%                   
& 69.11\%                         
& \cellcolor{lightblue}60.05\%   \\ 
\cmidrule(lr){1-11}
Ours                     
&   97.78\%                      
&   83.89\%                       
&   73.04\%                       
&   66.76\%                        
&   \multicolumn{1}{c|}{\cellcolor{lightblue}\best{80.37\%}}     
&   71.14\%                       
&   62.22\%                       
&   50.70\%                       
&   73.35\%                       
&   \cellcolor{lightblue}\best{64.35\%} \\ 
\bottomrule
\end{tabular}}
\end{table*}

\textbf{Datasets. }%
We evaluate our methods on three diverse classification tasks, each represented by a specific dataset. The first dataset is PACS~\cite{li2017deeper}, encompassing four distinct domains: Photo (P), Art (A), Cartoon (C), and Sketch (S). PACS comprises a total of seven classes. The second dataset, Office-Home~\cite{venkateswara2017deep}, involves four domains with 65 classes. The four domains are Art, Clipart, Product, and Real-World. 
The third dataset is IWildCam~\cite{koh2021wilds}, a real-world image classification dataset based on wild animal camera traps aroudn the world, where each camera represents a domain. It contains 243 training, 32 validation, and 48 test domains with a total of 182 classes. 

\noindent\textbf{FL Simulation.} 
Based on the framework provided by~\cite{bai2024benchmarking}, we simulate an FL system with a total of $N$ clients; in each training round, the server will randomly select $k\%$ of clients to participate, which is known as \textit{client sampling}. The data distribution of each client is simulated by \textit{domain-based client heterogeneity} by $\lambda$ level. We establish a rigorous FL scenario by adjusting the ratio of participating clients $k$ to total clients ($k$/$N$) and heterogeneity degree ($\lambda$). Unless otherwise noted, our default parameters are $N=100$ and $k=20\%$ for PACS/Office-Home, and $N=243$ and $k=10\%$ for IWildCam, with a heterogeneity level of $\lambda = 0.1$.
We set the number of communication iterations to 50 rounds for PACS/OfficeHome and 100 for IWildCam, and the number of local training epochs to 1.
The batch size is set to $32$ for all three datasets.

\noindent\textbf{Baselines.} 
We compare ours against five SOTA methods focusing on addressing domain generalization in FL, representative of three FedDG approaches in related works.
\subsection{Experimental Results}
\subsubsection{Comparison with State-of-the-Art.} First, we evaluate FedDG methods under different validation schemes by varying the domain(s) used for training. Specifically, we consider two evaluation methods: Leave-One-Domain-Out (LODO)~\cite{nguyen2022fedsr,huang2023rethinking} and Leave-Two-Domains-Out (LTDO)~\cite{bai2024benchmarking} with PACS and OfficeHome datasets. %

\textit{LTDO Scenarios.} 
Table~\ref{tab:ltdo_full} shows the final accuracy measurements with popular SOTA methods by the end of the FL process. We leave out two domains in each scheme alternately for validation and test accuracy.
These results suggest that FedSR is not well-suited in scenarios where each client contains a small amount of data (e.g., with a large number of clients), which aligns with observations in previous work~\cite{bai2024benchmarking}.
Other methods, such as FedGMA, FPL, and CCST, can somewhat generalize the FL model, 
but show degradation when training on domains less representative of the unseen domain, such as training on \textit{Photo} and testing on \textit{Cartoon}.
On the other hand, our method performs substantially better than other methods, confirming that it generalizes well and thus effectively boosts performance
on different unseen domains. Significantly, in the best case, our method outperforms with a gap of 3.64\% compared to the second-best method for the PACS dataset. 

\textit{LODO Scenarios.} Table~\ref{tab:lodo-full} presents the accuracy of compared methods with LODO experiments, i.e., we leave one domain out and use three domains for training. Generally, all methods achieve higher accuracy when more domains are used for training. Our method consistently performs best with PACS and OfficeHome datasets by achieving an average of 80.37\% and 64.35\% accuracy. Especially with domains such as Cartoon and Sketch of PACS dataset, we observe a large gap between our method and other baselines; our method outperforms the second-best method (i.e., FPL) by 4.63\% and 4.12\%, respectively.
\begin{figure*}[hbt]
\centering
\begin{minipage}[b]{.65\linewidth}
\includegraphics[width=1.0\textwidth]{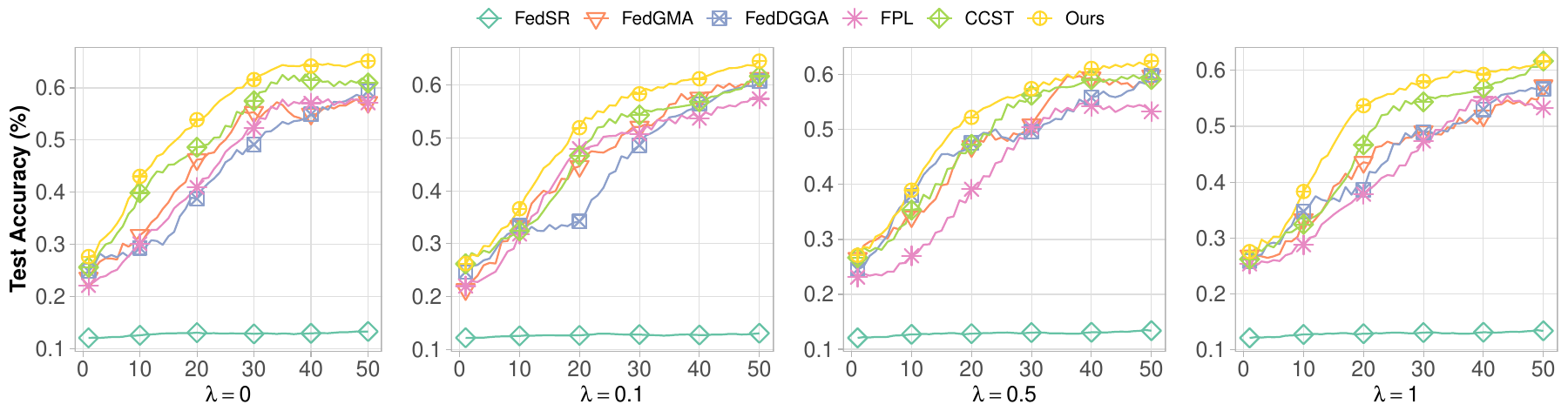}
\caption{Convergence curve when the training round increases on PACS's \textit{Sketch}, and the training domains are \textit{Art-Painting} and \textit{Cartoon}; decreasing domain heterogeneity from left to right. $\lambda = 0$ means domain separation while $\lambda = 1.0$ means homogeneous domain distribution.}
\vspace{-0.1cm}
\label{fig:pacs-heter}
\end{minipage}\hfill
\begin{minipage}[b]{.31\linewidth}
    \includegraphics[width=1.0\linewidth]{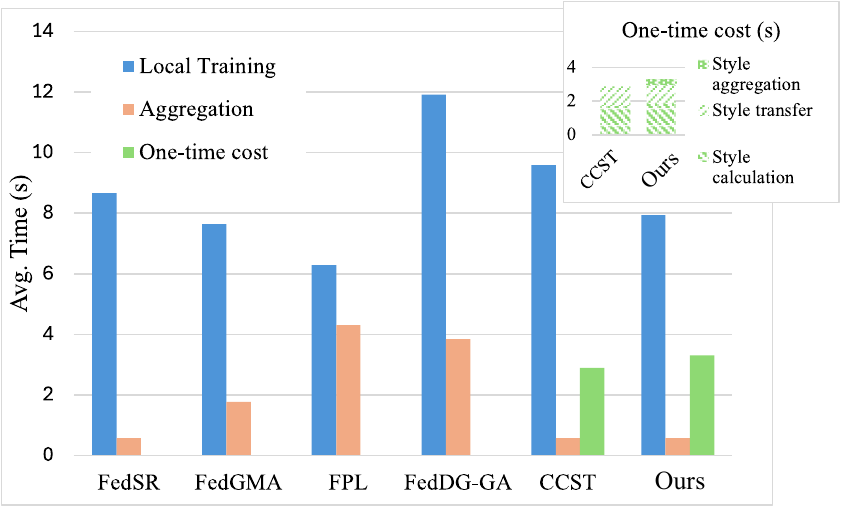}
    \caption{Computational overhead of FedDG methods.}%
    \label{fig:overhead}
\end{minipage}
\vspace{-0.2cm}
\end{figure*}

\subsubsection{Robustness Evaluation} We then study the
robustness of \thename{} w.r.t different domain heterogeneity, i.e., $\lambda$ and number of clients $K/N$.

\noindent\textbf{Impact of Domain Heterogeneity.} This experiment aims to study the effectiveness of different methods with different domain distribution settings. As shown in Figure.~\ref{fig:pacs-heter}, the larger the value of $\lambda$ is, the more heterogeneous domain distribution is among clients. We observe the accuracy when varying the domain heterogeneity using $\lambda = \left\{0.0, 0.1, 0.5, 1.0\right\}$ values. Figure.~\ref{fig:pacs-heter} shows that our method consistently outperforms other baselines under varied domain heterogeneity. Interestingly, our method also achieves the highest accuracy early in training, creating a substantial gap over other baselines.

Next, we evaluate FedDG methods on IWildCam, which is a challenging setting with 323 domains and 243 clients, and present the results in Table~\ref{tab:iwildcam-heterogeneity}. 
As observed in real-world data, the performance of FedDG methods significantly degrades as $\lambda$ decreases, consistent with findings by Bai et al.~\cite{bai2024benchmarking}. This degradation is particularly evident in baselines such as FedGMA, CCST, and FedDG-GA, where a lack of domain overlap ($\lambda=0.0$) leads local models to converge to biased solutions. Under these conditions, these baselines suffer performance drops ranging from 35\% to 50\%.
Notably, CCST, the second-best method on small-domain datasets, struggles remarkably in this challenging scenario. 
In contrast, our method exhibits robustness, experiencing only a 19\% performance drop in such scenarios. It achieves the best average accuracy across varying levels of heterogeneity on both validation and testing sets, demonstrating its effectiveness and stability under diverse conditions.
We argue that stability across varied levels of domain heterogeneity is a crucial requirement for practical FedDG methods.
\begin{figure}[thb]
    \centering
    \includegraphics[width=0.95\linewidth]{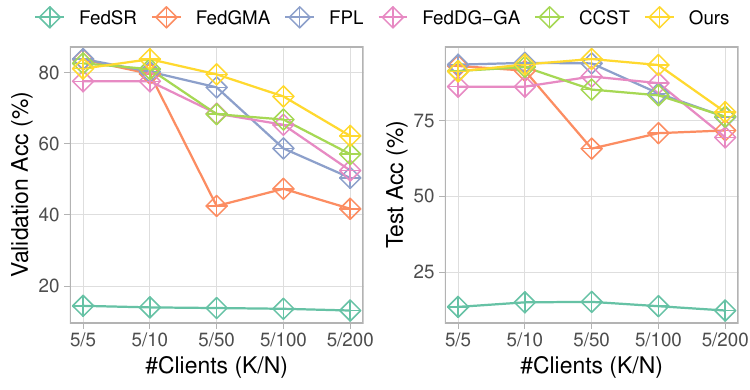}
    \captionof{figure}{Accuracy comparison  with different settings of selected clients (K) per total number of clients (N).}
    \label{fig:num_clients}
\end{figure}

\begin{figure}[tbh]
    \centering
    \begin{subfigure}{0.95\linewidth}
    \includegraphics[width=1.0\linewidth]{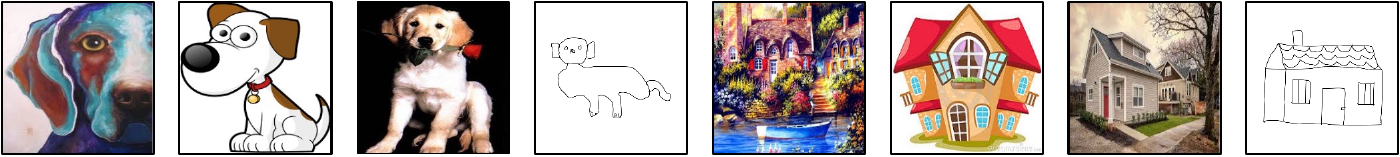}
    \caption{Original images were sampled from clients' training data.}
    \label{subfig:real-imgs}
    \end{subfigure}
    \begin{subfigure}{0.95\linewidth}
    \vspace{0.4cm}
    \includegraphics[width=1.0\linewidth]{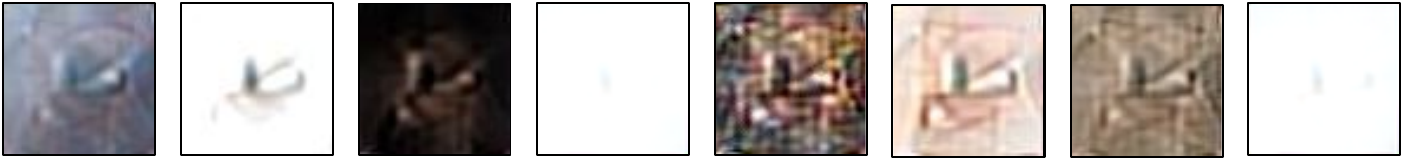}
     \caption{Reconstructed images using sample-level style vectors.}
      \label{subfig:fake-imgs-sample-level}
    \end{subfigure}
    \vspace{0.4cm}
    \begin{subfigure}{0.95\linewidth}
    \includegraphics[width=1.0\linewidth]{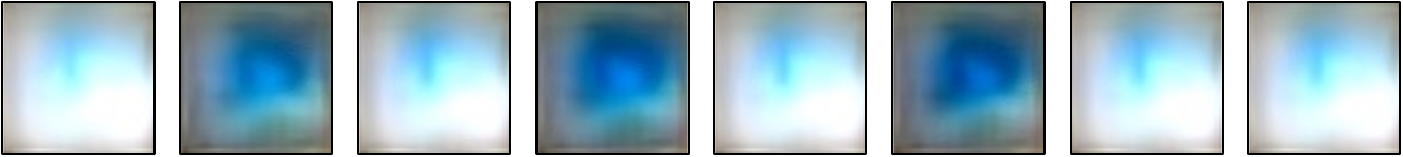}
     \caption{Reconstructed clients' images using client-level style vectors.}
     \label{subfig:fake-imgs-client-level}
    \end{subfigure}
    \vspace{-0.2cm}
    \caption{Image reconstruction from the GAN's models trained on the public dataset Tiny-ImageNet. The first row (Fig.~\ref{subfig:real-imgs}) is the real images sampled from clients' training data. The second row (Fig.~\ref{subfig:fake-imgs-sample-level}) shows the reconstructed images given the sample-level style vectors, i.e., vectors corresponding to each image in the first row. The last row (Fig.~\ref{subfig:fake-imgs-client-level}) presents images reconstructed given client's style vectors from different clients.}
    \vspace{-0.4cm}
\end{figure}

\begin{figure}[tbh!]
    \centering
    \begin{subfigure}{0.95\linewidth}
    \includegraphics[width=1.0\linewidth]{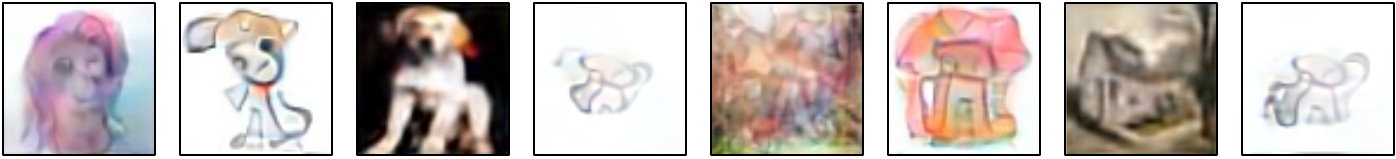}
    \caption{Reconstructed images using sample-level style vectors.}
    \label{subfig:ccst-imgs-02}
    \end{subfigure}
    \begin{subfigure}{0.95\linewidth}
    \vspace{0.4cm}
    \includegraphics[width=1.0\linewidth]{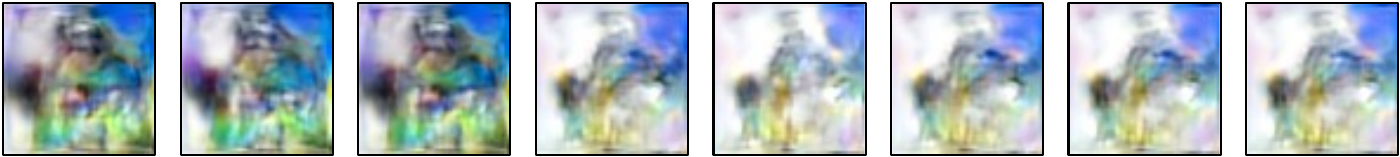}
    \caption{Reconstructed images using client-level style vectors.}
    \label{subfig:fisc-imgs-02}
    \end{subfigure}
    \caption{Image reconstruction from the GAN's models trained on the malicious client's dataset. The first row (Fig.~\ref{subfig:fake-imgs-sample-level}) shows the reconstructed images given the sample-level style vectors, i.e., vectors extracted from the original images. The last row (Fig.~\ref{subfig:fake-imgs-client-level}) presents images reconstructed given clients' style vectors from different clients.}
    \label{fig:malicious-reconstruct}
\end{figure}

\begin{figure}[thb!]
\centering
\vspace{-0.2cm}
\includegraphics[width=.9\linewidth]{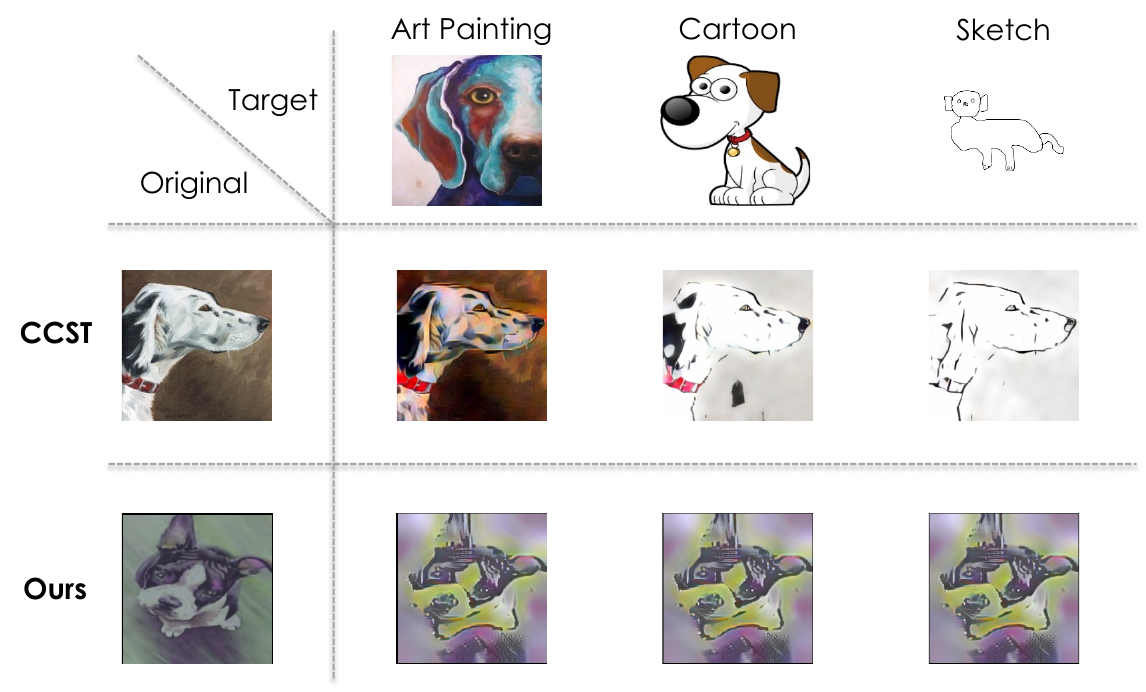}
\caption{Comparison of style-transferred images from ours and cross-client style transfer.}
\label{fig:illus-transfer}
\vspace{-0.5cm}
\end{figure}

\begin{table*}[bth]
\centering
\caption{Accuracy (\%) comparisons after 100 training rounds of the different methods with the IWildCam dataset, whose 243 training, 32 validation, and 48 test domains.}
\label{tab:iwildcam-heterogeneity}
\resizebox{0.85\linewidth}{!}{
\begin{tabular}{@{}lcccccccc@{}}
\toprule
& \multicolumn{4}{c}{Validation Accuracy}                                                                                                   & \multicolumn{4}{c}{Testing Accuracy}                                                                                                                              \\ 
\cmidrule(l){2-5} \cmidrule(l){6-9}
\multirow{-2}{*}{Methods} 
& \multicolumn{1}{c}{$\lambda = 0.0$} 
& \multicolumn{1}{c}{$\lambda = 0.1$} 
& \multicolumn{1}{c}{$\lambda = 1.0$} 
& \multicolumn{1}{c}{AVG}      
& \multicolumn{1}{c}{$\lambda = 0.0$} 
& \multicolumn{1}{c}{$\lambda = 0.1$} 
& \multicolumn{1}{c}{$\lambda = 1.0$} 
& \multicolumn{1}{c}{AVG} \\ \cmidrule(lr){1-9}
FedSR                     
& 3.12\%                             
& 32.97\%                             
& 5.57\%                             
& \multicolumn{1}{c|}{\cellcolor{lightblue}13.89\%} 
& 1.87\%                             
& 16.04\%                             
& 2.76\%                             
& \cellcolor{lightblue}6.89\%                 \\
FedGMA                    
& 39.26\%                             
& 71.17\%
& 74.65\%                             
& \multicolumn{1}{c|}{\cellcolor{lightblue}61.69\%} 
& 16.14\%                             
& 58.75\%
& 59.45\%                             
& \cellcolor{lightblue}44.87\%                 \\
FPL                       
& 45.61\%                             
& 70.45\%                             
& 72.44\%                             
& \multicolumn{1}{c|}{\cellcolor{lightblue}\second{62.83\%}} 
& 38.71\%                             
& 54.29\%                             
& 60.07\%                             
& \cellcolor{lightblue}51.02\%                 \\
FedDG-GA                  
& 37.33\%                             
& 67.98\%                             
& 71.72\%                             
& \multicolumn{1}{c|}{\cellcolor{lightblue}59.01\%} & 34.97\%                             
& 59.29\%                             
& 56.85\%                             
& \cellcolor{lightblue}50.37\%  \\ 
CCST                  
& 20.89\%                              
& 66.20\%                              
& 70.62\%                              
& \multicolumn{1}{c|}{\cellcolor{lightblue}52.57\%} 
& 38.01\%                              
& 56.58\%                              
& 58.95\%                              
& \cellcolor{lightblue}\second{51.18\%}                 \\ \midrule
\textbf{Ours}             
& 54.26\%                             
& 67.69\%                             
& 73.55\%                             
& \multicolumn{1}{c|}{\cellcolor{lightblue}\best{65.17\%}} 
& 44.03\%                             
& 59.34\%                             
& 60.58\%                             
& \cellcolor{lightblue}\best{54.65\%}                 \\ \bottomrule
\end{tabular}}
\vspace{-0.2cm}
\end{table*}

\noindent\textbf{Impact of Number of Clients.} The stability of different methods in federated learning (FL) depends on the number of clients, which can range from small (around ten) to large (hundreds or thousands). In many FL schemes, the server randomly selects $K$ participants from a total of $N$ in each training round, especially with many participants. We consider five cases: $N \in \{5, 10, 50, 100, 200\}$ and $K=5$, which correspond to 100\%, 50\%, 10\%, 5\%, 2.5\% clients participating in each round, respectively. 
We present the comparison of different methods in
Figure.~\ref{fig:num_clients}. The higher the ratio of $K/N$ is, the larger the amount of data participating in each training round is. The scenario of $5/5$ is the most similar setting to previous work, i.e., the number of clients is small, and all clients join each training round. 
While other methods, such as FPL and CCST, exhibit strong performance with a small number of clients, their efficacy diminishes with an increase in the total number of clients. In contrast, our method outperforms in terms of stability and efficiency.

\subsubsection{Scalability Study}In this section, we study the scalability of \thename{} under two dimensions, which are computational overhead and security preservation. 

\noindent\textbf{Computational Overhead.}
For the computation time measurement, we used consistent settings for all baselines, including the number of clients, each client's local data, and the indices of clients selected in each round. Then, for the local training time, we averaged the training time for all clients in each round for a fair comparison of each baseline. We break down the computational time into three sub-components, which are (i) local training at each client, (ii) aggregation time at the server, and (iii) the remaining one-time cost. 
As presented in Fig.~\ref{fig:overhead}, the average computation time \thename{} is comparable to or even smaller than other baselines. First, \thename{} does not introduce additional overhead during aggregation, which linearly increases concerning training rounds as in FedDGGA, FedGMA, or FPL. 
The overhead of average local training is comparable to other methods. 
The cost for interpolation style calculation is a one-time cost, happening before the training and only taking 3.3s, while the general local training takes 8.67s on average for all methods. 
This one-time cost does not increase linearly with the number of clients since all clients can conduct local style extraction simultaneously, and the clustering on the server is fast (only 0.3s). Therefore, \thename{} introduces low overhead when applied to large-scale systems.
We conclude that we \thename{} can ensure scalability on a large scale, where the number of clients can reach thousands.

\noindent\textbf{Security Analysis. } 
In our settings, there are two primary potential privacy breaches: 
(i) third-party reconstruction, where an external attacker or even the server compromises the shared style vectors to reconstruct private client datasets, and (ii) inter-client reconstruction, where a malicious client uses its data to reconstruct the images of other clients from their style vectors.
\begin{table}[tb!]
\centering
\vspace{-0.4cm}
\caption{Comparison metrics for the quality of reconstructed images from generative models.}
\label{tab:image-metrics}
\resizebox{0.95\linewidth}{!}{
\begin{tabular}{@{}lc|cc|cc@{}}
\toprule
\multirow{2}{*}{}    & \multirow{2}{*}{Domain} & \multicolumn{2}{c}{FID ($\uparrow$)}     & \multicolumn{2}{c}{Inception Score ($\downarrow$)} \\ \cmidrule(lr){3-4}  \cmidrule(lr){5-6}
                     &                         & Sample Style & Client Style & Sample Style     & Client Style     \\ \midrule
\multirow{4}{*}{\rotatebox{90}{Attack (i)}} & P                       & 203.24       & 422.75       & 0.29 $\pm$ 0.0025    & 0.16 $\pm$ 0.0022 \\
                     & A                       & 200.71       & 501.72       & 0.30 $\pm$ 0.0023  & 0.16 $\pm$ 0.0022  \\
                     & C                       & 224.50       & 475.19       & 0.30 $\pm$ 0.0016    & 0.16 $\pm$ 0.0022 \\
                     & S                       & 254.17       & 488.51       & 0.27 $\pm$ 0.0008    & 0.16 $\pm$ 0.0022 \\ \midrule
\multirow{4}{*}{\rotatebox{90}{Attack (ii)}} & P                       & 180.61       & 447.95       & 0.33 $\pm$ 0.0017      & 0.23 $\pm$ 0.0121      \\
                     & A                       & 193.51       & 483.31       & 0.35 $\pm$ 0.0016      & 0.23 $\pm$ 0.0121      \\
                     & C                       & 146.94       & 470.02       & 0.32 $\pm$ 0.0011      & 0.23 $\pm$ 0.0121      \\
                     & S                       & 203.43       & 502.01       & 0.28 $\pm$ 0.0008      & 0.23 $\pm$ 0.0121      \\ \bottomrule
\end{tabular}
}
\vspace{-0.5cm}
\end{table}

First, we conduct experiments to assess the potential of third-party/server reconstruction attacks, i.e., if an adversary compromises the style vectors and wants to reconstruct private training images from the shared style vectors using generative models~\cite{liu2020towards}. This analysis is crucial as local data includes sensitive medical records like X-rays and MRIs with identifiable details, where exposure risks stigma, discrimination, and job or insurance loss. We train the generator until the validation loss converges sufficiently, i.e., the loss is approximately reaching zero and barely changes. The best model with the highest validation average PSNR was selected. 
We visualized the reconstructed images for different clients in Fig.~\ref{subfig:fake-imgs-client-level}. 
The result shows that the reconstructed images are far different from the real images, i.e., compared to images from Fig.~\ref{subfig:real-imgs}. 
In addition, in our approach, each client uses only one vector to represent their data style, which leads to only one image that can be reconstructed for one client, i.e., no distinguishable images reflecting different classes are available. 
Indeed, compared to the case where the style of samples is allowed to be shared, adversaries can exploit it to reconstruct more distinguishable images as in Fig.~\ref{subfig:fake-imgs-sample-level}. We conclude that the design of \thename{} is more secure than other mechanisms allowing sample-level information exchange.

Regarding inter-client reconstruction, our method does not allow any cross-sharing mechanism between two arbitrary clients, which minimizes the privacy breaches between two clients compared to previous works~\cite{liu2021feddg,chen2023federated,park2024stablefdg}. We visualized the reconstructed images using sample-level styles and client-level styles (i.e., \thename{}) in Fig.~\ref{subfig:ccst-imgs-02} and Fig.~\ref{subfig:fisc-imgs-02}, respectively. From the results, this attack is extreme because if the adversary can acquire a sample-level style vector (i.e., single style as in CCST), they can reconstruct the content of the image (cf. Fig.~\ref{subfig:ccst-imgs-02}). However, with \thename{} method, the reconstructed images do not contain content information of the private data. 
We provide quantitative results in Table~\ref{tab:image-metrics} to further support our conclusion, where the images generated using \thename{}'s style vectors are much less informative and have lower quality than strategies using sample-level style.
We conclude that reconstructing a client's data solely from its style vector is nontrivial and requires specialized methods tailored to the data reconstruction tasks given style vectors utilized in our approach.

Lastly, we present the visualization of style-transferred images generated by our method and CCST, a baseline that uses style transfer for data augmentation. 
As shown in Fig.~\ref{fig:illus-transfer}, the images produced by \thename{} across different target domains are visually indistinguishable. In contrast, the images generated by CCST exhibit distinguishable differences among domains, as sample-level styles are communicated. This results in transferred images that closely resemble the private datasets of other clients, unlike those generated by our method. 
By comparing the images, we conclude that interpolation-style transfer offers stronger privacy protection than cross-local style transfer, as demonstrated by the comparison with CCST~\cite{chen2023federated}.

In conclusion, 
the local style statistics component of \thename{} does not introduce additional security risks; instead, it mitigates attack threats compared to the sample-level and cross-client sharing employed in other works.

\subsubsection{Ablation Study}

\begin{table}[tb!]
\centering
\vspace{-0.4cm}
\captionof{table}{Performance of \thename{} with different components present -- removed components (\xmark) / retained (\cmark).}
\label{tab:ablation-study}
\resizebox{0.95\linewidth}{!}{

\begin{tabular}{@{}lccccc@{}}
\toprule
\multirow{2}{*}{Methods} & \multicolumn{3}{c}{Components}                                                                                                                                                                                   & \multicolumn{2}{c}{Performance} \\ \cmidrule(lr){2-4} \cmidrule(lr){5-6} 
 & \multicolumn{1}{c}{\begin{tabular}[c]{@{}c@{}}Local \\ Clustering\end{tabular}} 
 & \begin{tabular}[c]{@{}c@{}}Global \\ Clustering\end{tabular} 
 & \begin{tabular}[c]{@{}c@{}}Contrastive \\ Learning\end{tabular} 
  
 & \multicolumn{1}{c}{\begin{tabular}[c]{@{}c@{}} Validation \\Accuracy
 \end{tabular}}
 & 
 \multicolumn{1}{c}{\begin{tabular}[c]{@{}c@{}} Test \\Accuracy \end{tabular}}      \\ \midrule
\thename{}-v1               
& \multicolumn{1}{c}{\xmark}                                       
& \cmark
& \multicolumn{1}{c|}{\cmark}           
& 72.90\%                
& 92.22\%                       
               \\
\thename{}-v2               
& \multicolumn{1}{c}{\cmark}                                       
& \xmark
& \multicolumn{1}{c|}{\cmark}                            
& 72.80\%       
& 92.18\%
         
               \\
\thename{}-v3               
& \multicolumn{1}{c}{\cmark}                                       
& \cmark           
& \multicolumn{1}{c|}{\xmark}
& 64.89\%
& 85.69\%                    \\
\thename{}-v4               
& \multicolumn{1}{c}{\xmark}                                       
& \xmark     
& \multicolumn{1}{c|}{\cmark}                                          
& 59.42\%                                          
& 82.33\%                   \\ \cmidrule{1-6}
\thename{}-v5               
& \multicolumn{1}{c}{\cmark}                                       
& \cmark     
& \multicolumn{1}{c|}{\cmark}                                         
& \best{73.63\%}                       
& \best{93.05\%}                    \\ \bottomrule
\end{tabular}}
\vspace{-0.4cm}
\end{table}

In Table~\ref{tab:ablation-study}, we conduct an ablation study to investigate the impacts of the main components proposed in our \thename{} approach. We gradually remove each component from \thename{} architecture to construct four incomplete versions of \thename{}.
In the \thename{}-v1 and \thename{}-v2 versions, the \xmark{} cells indicate that simple averaging was used to calculate the local and global styles, rather than the clustering methods described earlier. The results demonstrate that clustering at both the local and global levels is essential in mitigating the bias introduced by domain heterogeneity, proving to be more effective than simple averaging. The performance limitations of the simple averaging approach underscore the necessity for clustering, with FINCH outperforming simple averaging in both versions.
The contrastive learning component in \thename{}-v3 is identified as the second-most critical element. Excluding this component leads to a significant 9\% drop in accuracy, even when style-transferred data is added to the training. This result underscores the importance of the carefully designed combination of interpolation style and contrastive learning. However, it is important to note that the effectiveness of the contrastive learning component hinges on the use of interpolation style-transferred data. This is evident from the comparison between \thename{}-v4 and \thename{}-v5, where \thename{}-v4—relying on standard contrastive learning with augmentation and close samples as positive anchors—fails to address domain shifts effectively under this scenario, unlike \thename{}-v5, which utilizes interpolation style-transferred data.
In conclusion, the \thename{}-v5 version, which incorporates all components, demonstrates superior performance in handling unseen domain accuracy, reinforcing the value of the well-designed components of \thename{}.

\section{Conclusions}
\label{sec:conclusion}
This work presents \thename{}, a domain generalization method in FL. \thename{} extracts an unbiased interpolative style from all clients to facilitate style transfer. Each client then performs style transfer using this information through contrastive learning, aiming to enhance generalizability and mitigate local bias. Consequently, the aggregated model achieves improved accuracy for unseen domains. Extensive experiments demonstrate \thename{}'s improved performance over existing FedDG baselines across various settings, from small to large domain datasets, and in scenarios with high domain heterogeneity. In addition, using an interpolation style helps enhance the local privacy of clients compared to a cross-client style's information exchange mechanism.

\bibliographystyle{IEEEtran}
\bibliography{IEEEabrv,main}

\end{document}